\documentclass{svmult}

\usepackage{mathptmx}
\usepackage{helvet}
\usepackage{courier}
\usepackage{amsmath,amsfonts}
\usepackage{graphicx}
\usepackage[bottom]{footmisc}
\usepackage{enumerate}
\usepackage[dvips]{floatflt}

\begin{document}
\title*{Solution regions in the parameter space of a 3-RRR decoupled robot for a prescribed workspace}
\author{D. Chablat \and G. Moroz \and V. Arakelian \and S. Briot \and P. Wenger}
\institute{%
  D. Chablat \at
  Institut de Recherche en Communications et Cybernétique de Nantes,\\
  \email{damien.chablat@irccyn.ec-nantes.fr}
} 

\titlerunning{Solution regions in the parameter space of a 3-RRR decoupled robot\ldots}

\maketitle
\abstract{This paper proposes a new design method to determine the feasible set of parameters of translational or position/orientation decoupled parallel robots for a prescribed singularity-free workspace of regular shape. The suggested method uses Groebner bases to define the singularities and the cylindrical algebraic decomposition to characterize the set of parameters. It makes it possible to generate all the robot designs. A 3-RRR decoupled robot is used to validate the proposed design method.}

\section{Introduction}
Parallel robots are attractive for various reasons but one has to cope with their singularities. There exists three main ways of coping with singularities, which have their own merits. A first approach consists in eliminating the singularities at the design stage by properly determining the kinematic architecture, the geometric parameters and the joint limits \cite{Arsenault:2004,Wenger:1999}. This approach is difficult to apply in general and restricts the design possibilities but it is safe. A second approach is the determination of the singularity-free regions in the workspace \cite{Merlet:1998,Li:2006}. This solution does not involve a priori design restrictions but it may be difficult to determine safe regions that are sufficiently large. Finally, a third way consists in planning singularity-free trajectories in the manipulator workspace \cite{Sen:2003}. In this paper, the first approach is used. Designing a parallel robot that will operate in a singularity-free workspace is a first requirement but the designer often needs to optimize the robot as function of various criteria \cite{McCarthy:2010}. Our goal is to generate the set of geometric parameters for a given singularity-free workspace. The resulting solution regions in the parameter space are of primary interest for the designer. 
Accordingly, this paper proposes a new design method to determine these solution regions. This method holds for parallel translational robots and for parallel robots with position/orientation decoupled architecture. Groebner bases are used to define the singularities and Cylindrical algebraic decomposition is applied to characterize the set of design parameters. 
The paper is organized as follows. Section 2 introduces the design method to generate the solution regions in the parameter space for a prescribed workspace of regular shape. Then, Section 3 applies this method to a 3-RRR planar parallel robot with position/orientation decoupled architecture.
\section{Design method}
\subsection{Definition of the prescribed regular workspace}
A robot should have sufficiently large, regular workspace with no singularity inside \cite{Chablat:2004}. For planar (resp. spatial) translational robots, a regular workspace can be defined by a circle, a square or a rectangle (resp.  a cylinder, a cube or parallelepiped).
A circle, a cylinder or a sphere can be modeled with one single algebraic equation. A rectangle or a parallelepiped can be defined with a set of linear equations. It can be approximated using a Lamé curve (resp. surface). This approximation makes it possible to handle only one equation, thus simplifying the problem resolution as will be shown further. In the rest of the paper, the problem is formulated in the plane for practical reasons.
A Lam\'e curve based workspace ${\cal W}_L$  can be defined by the following boundary algebraic equation:
\begin{equation}
{\cal W}_L: \bigl(\tfrac{x-x_c}{l_x/2}\bigr)^n+\bigl(\tfrac{y-y_c}{l_y/2}\bigr)^n=1
\end{equation}
$l_x$ and $l_y$ being the edge lengths of the desired rectangle, $n$ being a strictly positive integer. For the purpose of this paper, $n=4$ and $l_x=l_y=4$.
A rectangle based workspace can be modeled by four parametric lines, noted  $\overline{{\cal W}}_{Ci}$
\begin{equation}
\overline{{\cal W}}_{Ci}: \left\{ 
\begin{array}{c}
	x=P_{(i)x}t+ P_{(i+1)x}(1-t)\\
	y=P_{(i)y}t+ P_{(i+1)y}(1-t)
\end{array} \right. {\rm with~} t \in [0~1], i=1, 2, 3, 4
\end{equation}
$P_{(i)x}=x_c\pm l_x / 2 \quad P_{(i)y}=y_c\pm l_y / 2$ where $P_i$ denote the rectangle vertices. 
For position/orientation decoupled robot architectures, the regular workspace is defined using the same approach for the translational module and the orientation module is considered separately, as it will be shown in the next section. 
\subsection{Method to generate the solution regions in the parameter space}
The problem can be stated as follows:  find the regions in the parameter space where the boundaries $\overline{{\cal W}}$   of the workspace ${\cal W}$  have no intersection with the serial and parallel singularities loci $\delta_i$, namely:
\begin{equation}
{\cal P}: [a_1 \ldots a_n] / \delta_i \cap \overline{{\cal W}}= \emptyset, a_j>0, j=1, \ldots, n 
\end{equation}
where $[a_1 \ldots a_n]$  are the set of design parameters. This approach stands if and only the singularity curves or points are never fully included in the prescribed region. In order to find the design parameters for which the intersection is empty, the design parameters will be sorted according to the number of intersections between the singularities and  $\overline{{\cal W}}$.
It is then necessary to decompose the design parameter space into cells  $C_1, \ldots, C_k$, such that: 
(a)  $C_i$ is an open connected subset of the design parameter space;
(b) for all design parameter values in  $C_i$, the design parameter space has a constant number of solutions.
This analysis is done in 3 steps \cite{Lazard:2007}:
\begin{enumerate}[(a)]
 	\item computation of a subset of the joint space (workspace, resp.) where the number of solutions changes: the {\it Discriminant Variety};
 	\item description of the complementary of the discriminant variety in connected cells: the {\it Generic Cylindrical Algebraic Decomposition};
 	\item connecting the cells that belong to the same connected component of the complementary of the discriminant variety: {\it interval comparisons}.
\end{enumerate}
\begin{floatingfigure}{35mm}
 				\includegraphics[height=0.15\textheight]{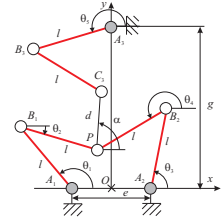}
        \caption{The 3-RRR decoupled parallel robot under study}
        \protect\label{Figure_01}
\end{floatingfigure}
The results are sets of regions with the same number of intersections between $\delta_i$  and  $\overline{{\cal W}}$. These three steps were integrated in a single function in the Siropa Library implemented in Maple (Moroz, 2010). For the purpose of this study, only the solutions with zero intersections are considered.
When a decoupled robot is analyzed, problem $\cal P$  is first treated for a prescribed workspace and a slightly modified problem ${\cal P}'$ is then treated, in which the set of design parameters include the orientation parameters. This approach is illustrated in the next section.
\section{Application to a 3-RRR decoupled parallel robot}
The robot under study is a planar 3-RRR robot with a modified mobile platform design \cite{Arakelian:2011} (Fig.~\ref{Figure_01}), thus decoupling the position and the orientation of the platform \cite{Gogu:2009}. It is assumed to have three identical legs.
The loop $(A_1, B_1, P, B_2, A_2)$ corresponds to a five-bar robot that defines the position of point $P$ and the leg $(A_3, B_3, C_3)$ adjusts the orientation according to the position. If position of point $P$ is given, this third leg is equivalent to a four-bar linkage. For this 3-RRR robot, thus, the problem can be split into two parts: (i) design of the five-bar robot (the translational module) so that the end-effector can move in a prescribed singularity-free workspace and (ii) design of the third leg (the four-bar linkages $(A_3, B_3, C_3, P)$) so that the platform can be oriented within desired bounds throughout the prescribed workspace.
\subsection{Translational module: five-bar robot}
The constraint equations   of the five-bar robot are defined as:
\begin{equation}
{\cal C}_i: 
\left\{
\begin{array}{c}
x -l \cos(\theta_1)-l \cos(\theta_2)+e/2=0 \quad 
y -l \sin(\theta_1)-l \sin(\theta_2)=0\\
x -l \cos(\theta_3)-l \cos(\theta_4)-e/2=0 \quad
y -l \sin(\theta_3)-l \sin(\theta_4)=0\\
\end{array}
\right.
\end{equation}
where $\left\|{\bf A}_1{\bf B}_1\right\|= \left\|{\bf A}_2{\bf B}_2\right\|= \left\|{\bf B}_1{\bf P}\right\|= \left\|{\bf B}_2{\bf P}\right\|=$   and  $\left\|{\bf A}_1{\bf A}_2\right\|=e$. The differentiation of the relation between the input variables $\bf q$   and the output variables $\bf X$  with respect to time leads to the velocity model $
{\bf At} + {\bf B\dot q}=0
$ where $\bf A$ and $\bf B$  are  $n \times n$ Jacobian matrices, $t$ is the platform twist and $\dot{\bf q}$ is the vector of joint rates.
The roots of the determinant of $\bf A$  and $\bf B$  define the parallel and serial singularities, respectively. The first ones are directly characterized in the workspace and the second ones have to be projected from the joint space onto the workspace. The singularities are calculated using Groebner bases \cite{Lazard:2007} as in \cite{Moroz:2010}. 

The parallel singularities can be factored into a sextic, denoted $\delta_{p1}$, and two quadratic polynomial equations, denoted $\delta_{p2}$  and  $\delta_{p3}$
\begin{eqnarray}
 \delta_{p1}&:&
16(y^6+x^6)+8(e^2y^4-e^2x^4)+48(y^4x^2+y^2x^4)+e^4y^2+e^4x^2-16l^2e^2y^2=0  \nonumber	\\
 \delta_{p2}&:& x^2+\left(y-\frac{1}{2}\sqrt{4l^2-e^2}\right)^2-l^2=0 \quad
 \delta_{p3}: x^2+\left(y+\frac{1}{2}\sqrt{4l^2-e^2}\right)^2-l^2=0 \nonumber
\end{eqnarray}
The serial singularities are two quadratic equations
\begin{equation*}
 \delta_{s1}:  (2x+e)^2+4y^2-16l^2=0	\quad  \delta_{s2}:  (2x-e)^2+4y^2-16l^2=0 
\end{equation*}

Due to the symmetry of the robot with respect to y-axis, the design parameters are restricted to $(l~f)$  i.e. the size of the legs and the distance from axis x to the geometric center of the robot's workspace $\cal W$, respectively. Parameter $e$ is set to 1 to have a two dimensional representation of the solution regions. For robots with two degrees of freedom, the intersection of the boundaries of $\cal W$  and the singularities is generically a finite set of points. Thus, as mentioned in §2.2, the singularity curves or points are never fully included in the prescribed region.

\begin{figure}[hbtp]
     \begin{center}
 				\includegraphics[height=0.15\textheight]{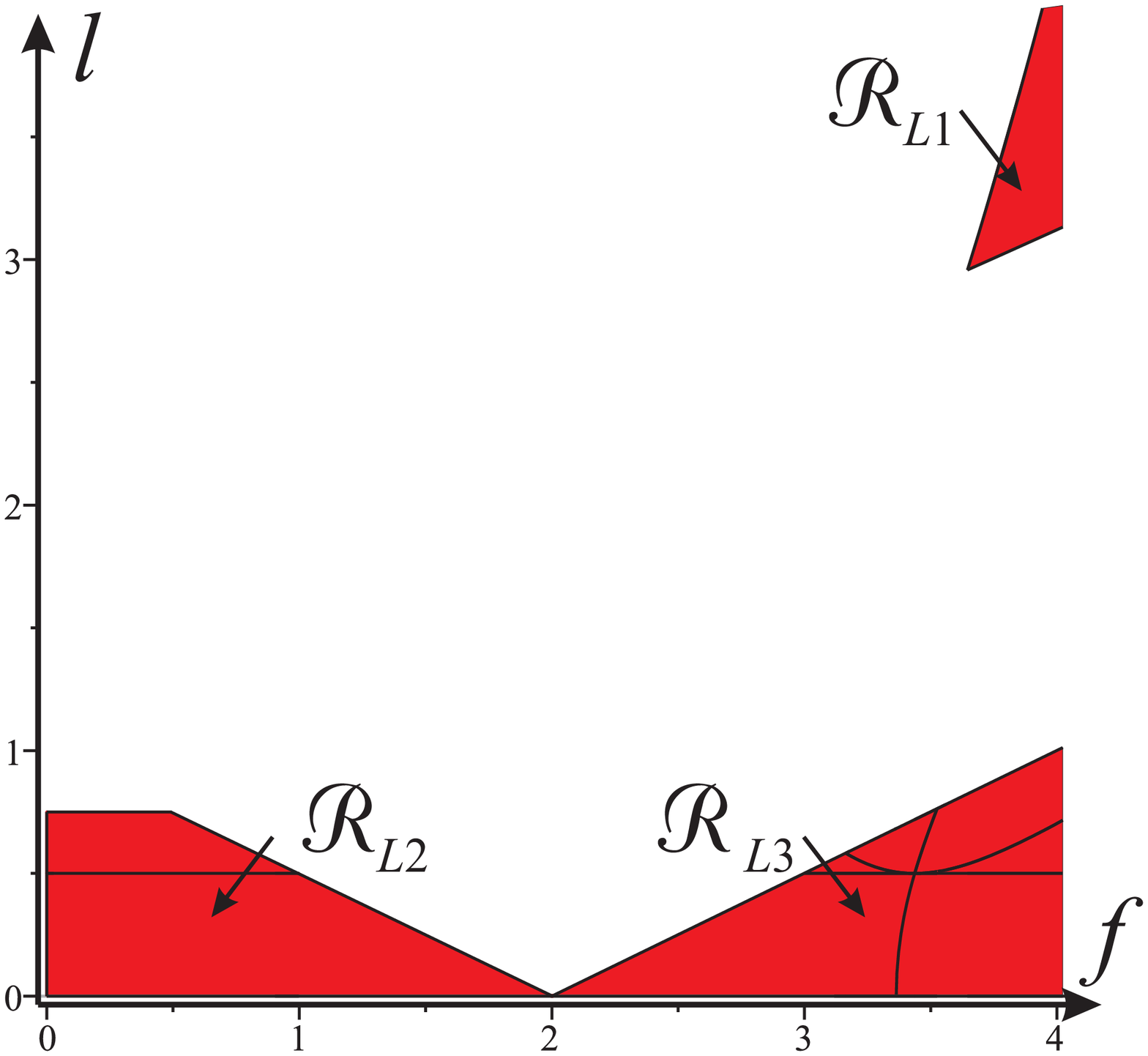}{\small(a)}
 				\includegraphics[height=0.15\textheight]{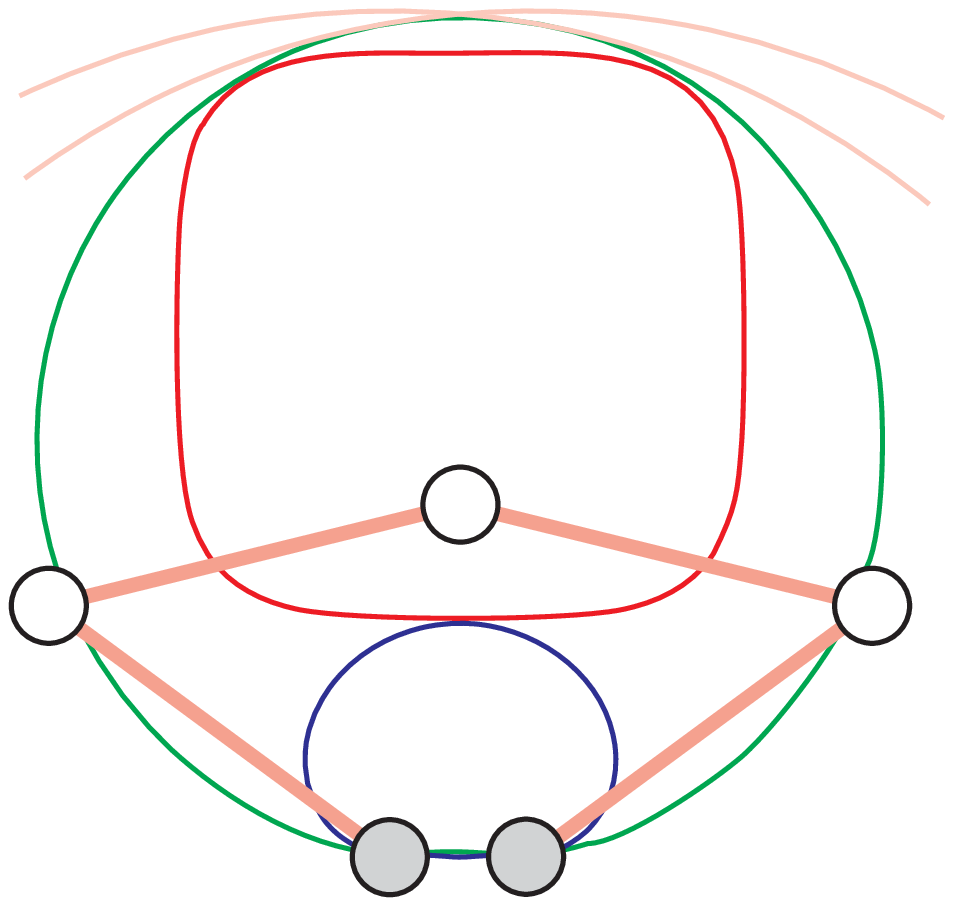}{\small(b)}
 				\includegraphics[height=0.15\textheight]{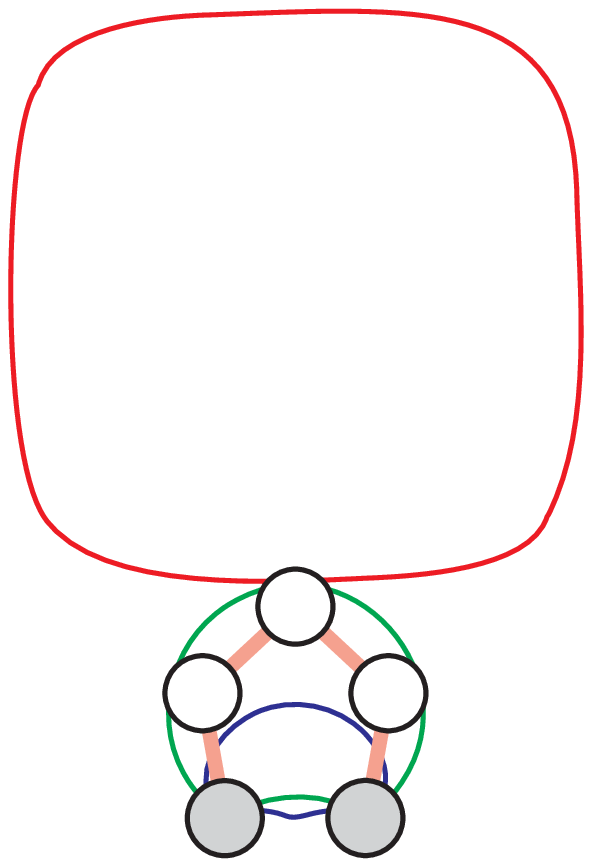}{\small(c)}
        \caption{(a) Solution regions  ${\cal R}_{L1}$, ${\cal R}_{L2}$  and ${\cal R}_{L3}$ of problem ${\cal P}_L$  and five-bar robot design when (b) $f=3.7$, $l=3$ (c) $f=3.7$, $l=0.9$}
        \protect\label{Figure_02}
     \end{center}
\end{figure}

\underline{Lamé curve based workspace:} the problem to be solved is:
\begin{equation*}
{\cal P}_{L}:[f~l]/{\cal S}_{p1} \cap	{\cal S}_{p2} \cap {\cal S}_{p3} \cap {\cal S}_{s1} \cap {\cal S}_{s2} \cap \overline{\cal W} =\emptyset, f>0, l>0	
\end{equation*}
Only the solutions with zero intersections are kept. Fig.~\ref{Figure_02} depicts the three solution regions obtained ${\cal R}_{L1}$, ${\cal R}_{L2}$  and  ${\cal R}_{L3}$, i.e. the parameter sets for which the prescribed workspace is singularity-free.

It turns out that in  ${\cal R}_{L1}$, ${\cal W}_L$  is inside the workspace (Fig.~\ref{Figure_02}b). Conversely, in  ${\cal R}_{L2}$ and  ${\cal R}_{L3}$, ${\cal W}_L$  is outside the workspace (Fig.~\ref{Figure_02}c). Thus the only feasible region is  ${\cal R}_{L1}$.  A feasible solution should not be taken on the boundary of ${\cal R}_{L1}$ since any solution on the boundary could touch a singularity curve. Fig.~\ref{Figure_02}b shows a solution near the boundary of  ${\cal R}_{L1}$. 

\begin{figure}[hbtp]
     \begin{center}
 				\includegraphics[height=0.15\textheight]{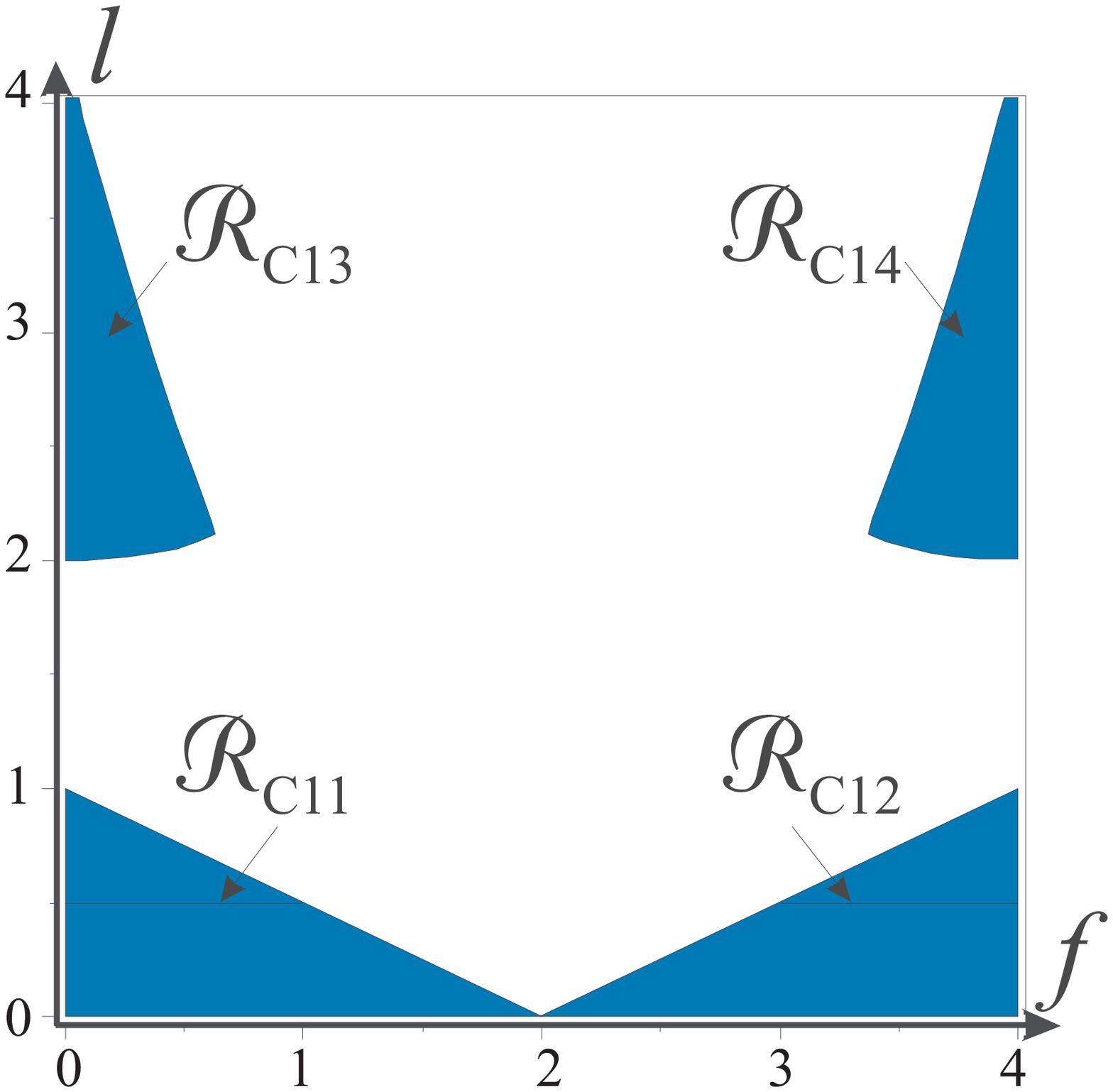}{\small(a)}
 				\includegraphics[height=0.15\textheight]{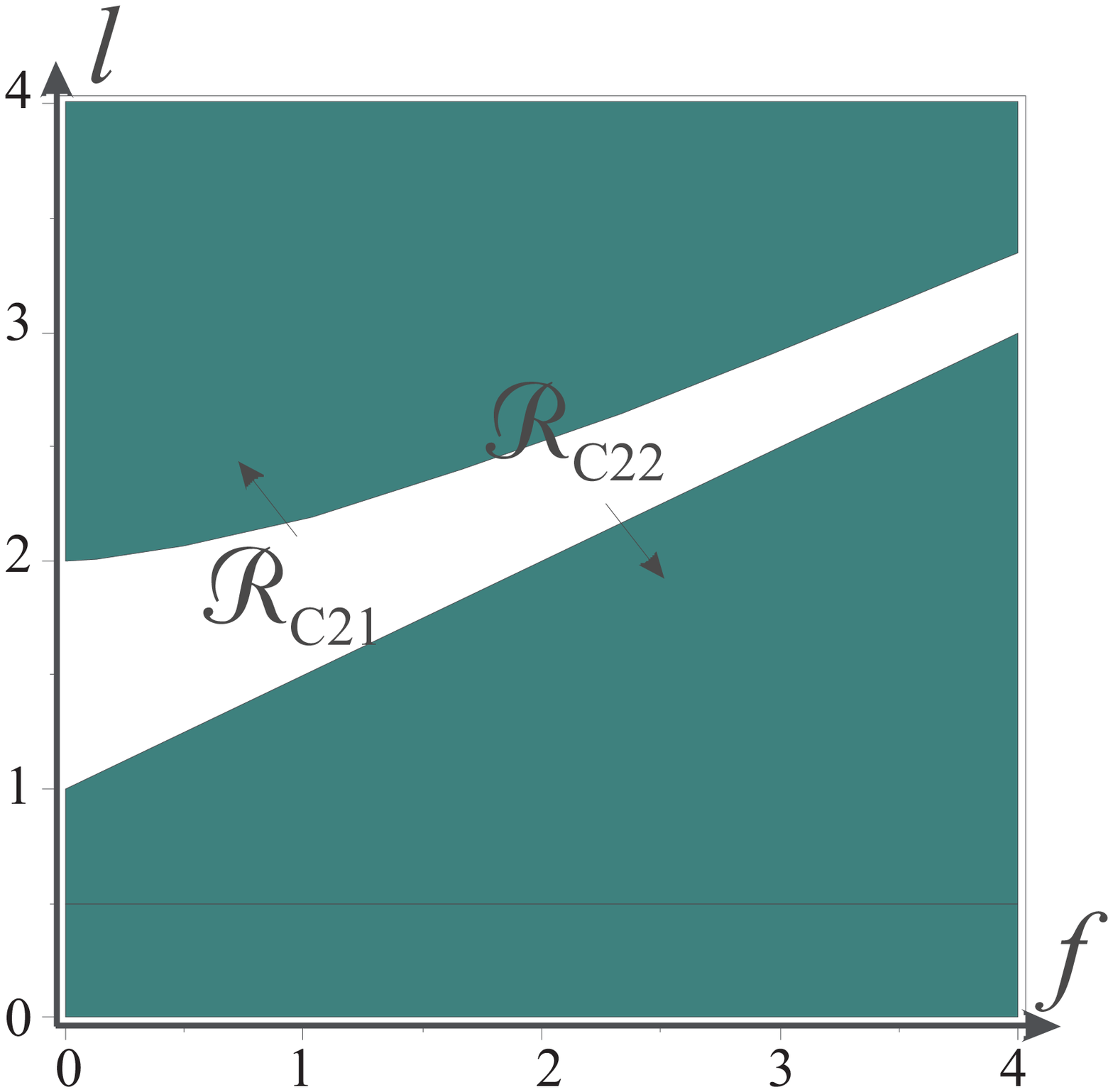}{\small(b)} 
 				\includegraphics[height=0.15\textheight]{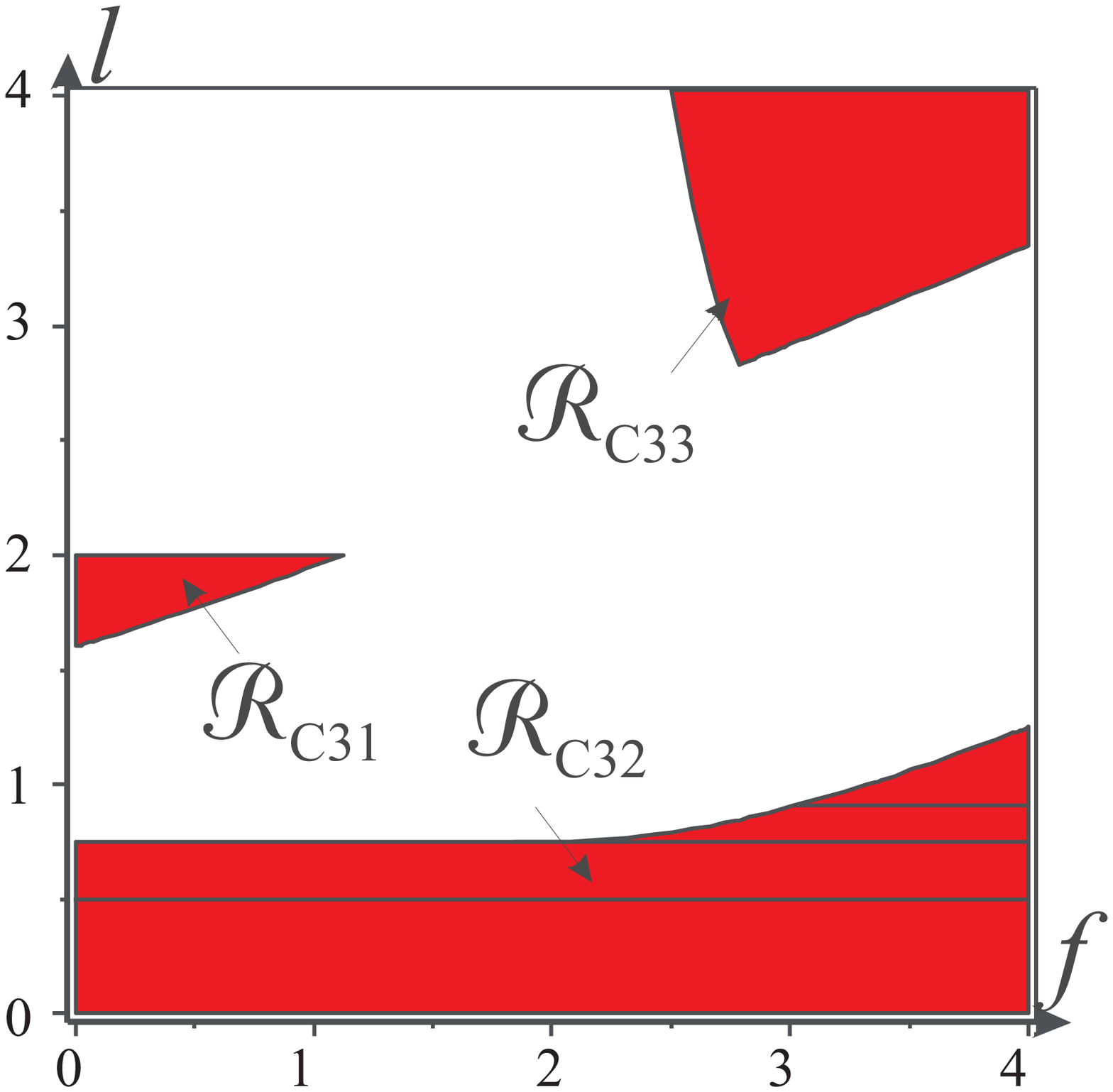}{\small(c)}
        \caption{Solution regions for problems (a) ${\cal P}_{C1}$, (b) ${\cal P}_{C2}$, (c) ${\cal P}_{C3}$ and (d) intersection regions ${\cal R}_{Cf1}$, ${\cal R}_{Cf2}$  and ${\cal R}_{Cf3}$}
        \protect\label{Figure_03}
     \end{center}
\end{figure}

\begin{figure}[hbtp]
    \begin{tabular}{cc}
       \begin{minipage}[t]{40 mm}
           \includegraphics[height=0.15\textheight]{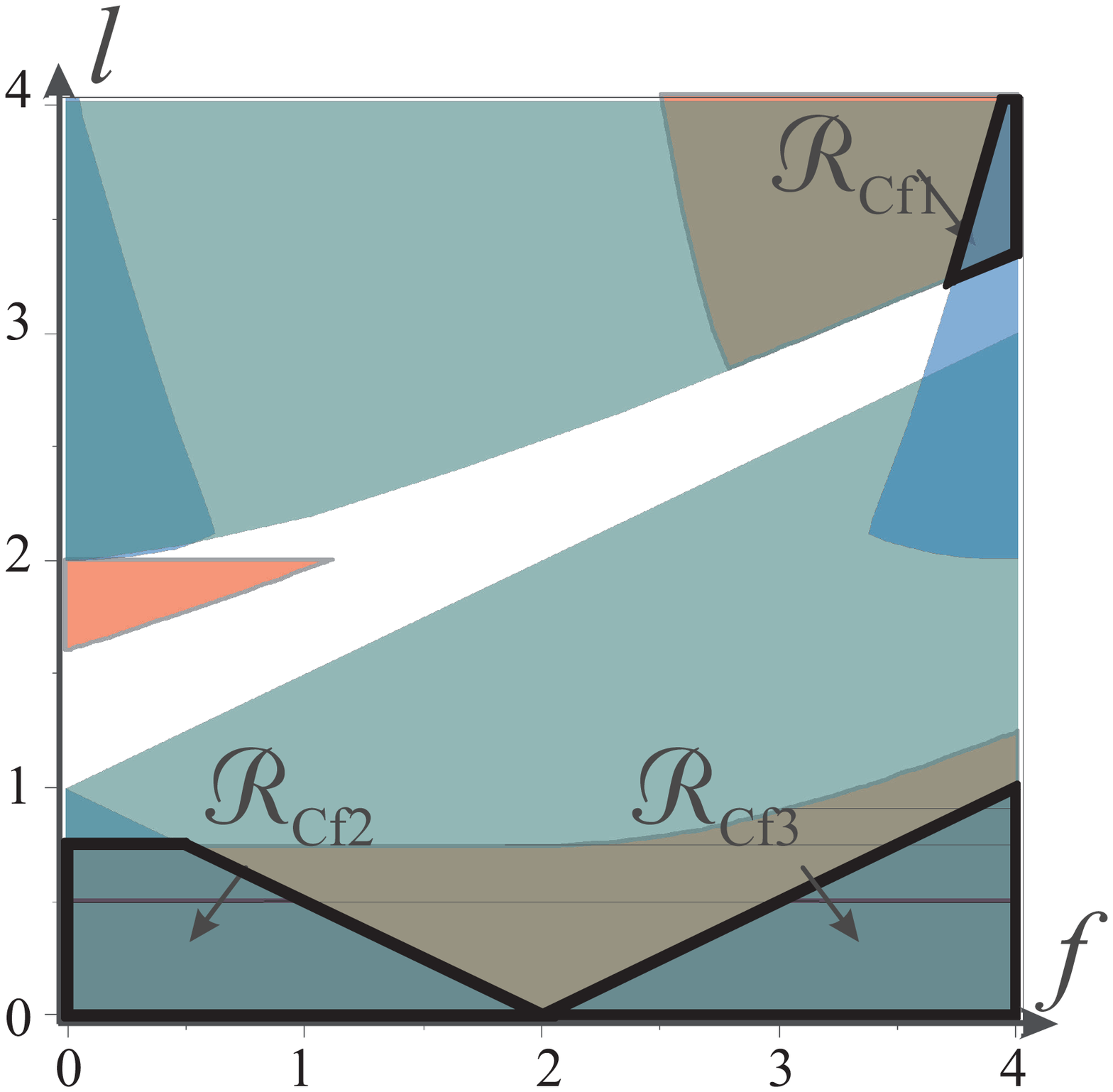}
           \caption{Intersection regions ${\cal R}_{Cf1}$, ${\cal R}_{Cf2}$  and ${\cal R}_{Cf3}$}
           \protect\label{Figure_03d}
       \end{minipage} &
       \begin{minipage}[t]{120 mm}
 				\includegraphics[height=0.15\textheight]{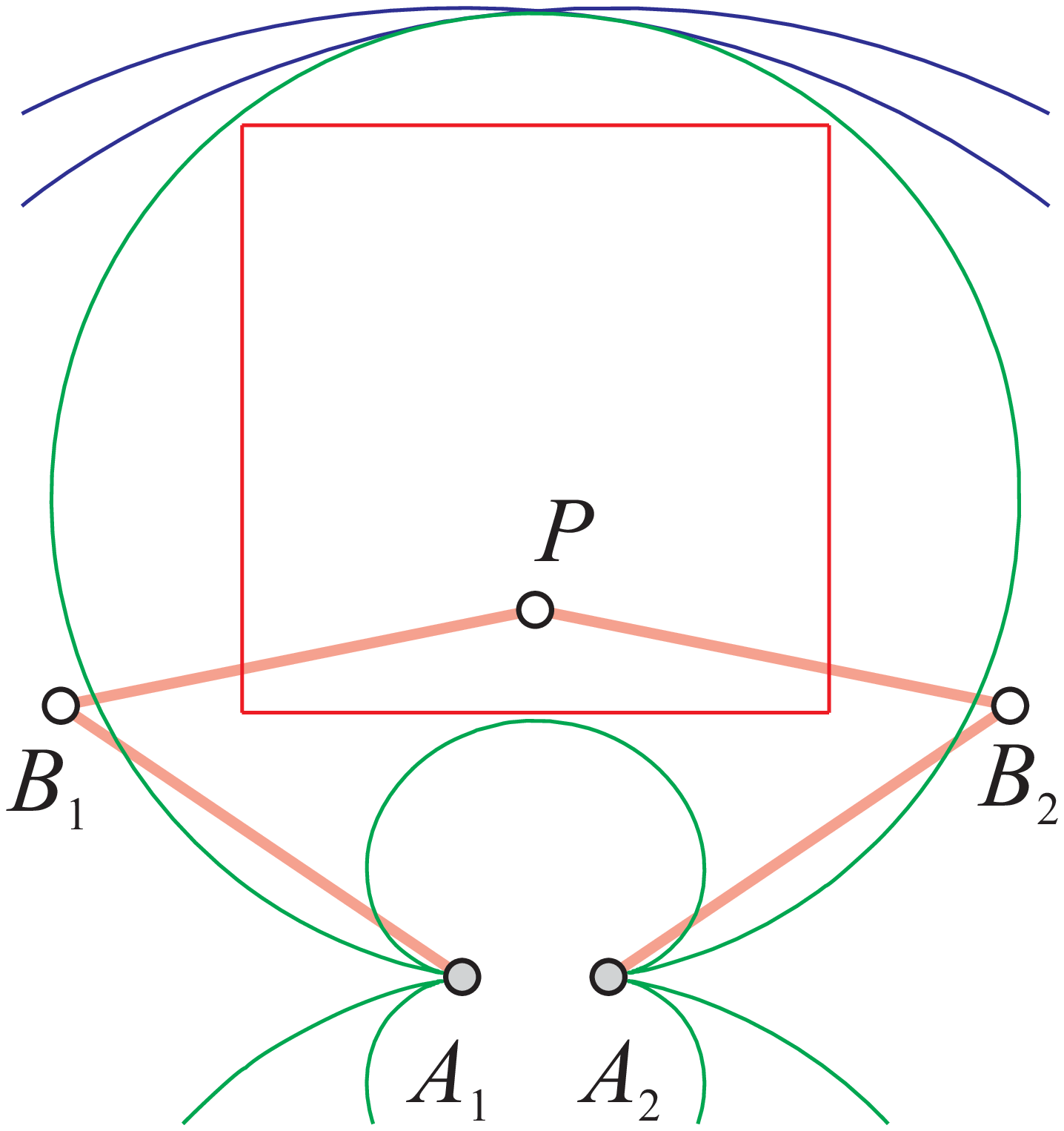}{\small(a)}
 				\includegraphics[height=0.15\textheight]{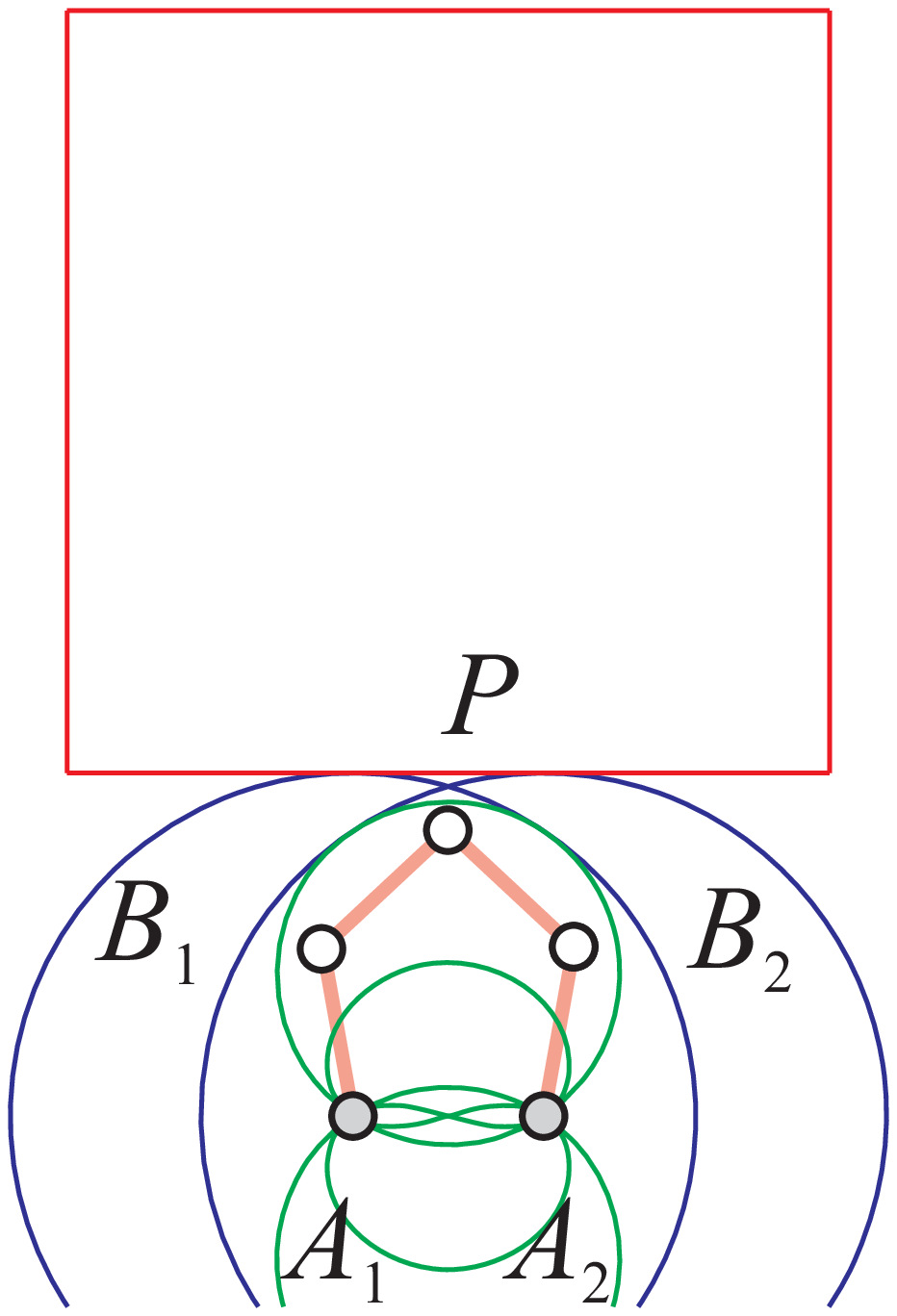}{\small(b)}         
 				\caption{A five-bar solution robot when $f=3.8$, $l=3.3$ from \goodbreak ${\cal R}_{Cf1}$  (a) and when $f=3.8$, $l=0.9$ from  ${\cal R}_{Cf2}$(b)}
        \protect\label{Figure_04}
      \end{minipage}
    \end{tabular}
\end{figure}

\underline{Square based workspace:} in this case, four separate problems need to be solved:
\begin{equation*}
{\cal P}_{Ci}: [f~l]/ {\cal S}_{p1} \cap {\cal S}_{p2} \cap {\cal S}_{p3} \cap {\cal S}_{s1} \cap {\cal S}_{s2} \cap \overline{\cal W}_{Ci}=\emptyset, f>0, l>0, t \in[0,1], i=1, \ldots, 4
\end{equation*}
where   are the parametric equations defining the boundaries of the square. Only the solutions with zero intersections are kept.
Due to the symmetry of the square with respect to the y-axis, ${\cal P}_{C3}$  and ${\cal P}_{C4}$  yield the same regions in the design parameters space. Fig.~\ref{Figure_03} depicts (a) four connected solution regions for problem ${\cal P}_{C1}$, (b) two solution regions for ${\cal P}_{C2}$  and (c) three solution regions for  ${\cal P}_{C3}$.
As compared to the Lam\'e curve based workspace, there is an additional step here: the final regions must be obtained by intersecting all these regions, thus yielding the three regions ${\cal R}_{Cf13}$, ${\cal R}_{Cf2}$  and ${\cal R}_{Cf3}$  as shown in Fig.~\ref{Figure_03d}. As expected, the solution regions obtained are similar to those associated with a Lam\'e curve (Fig.\ref{Figure_02}) and only ${\cal R}_{Cf1}$  is solution to the problem for the same reasons. Fig.~\ref{Figure_04}a shows a solution near the boundary of  ${\cal R}_{Cf3}$.
\subsection{Orientation module: four-bar linkages}
One of the two base points of the four-bar linkages is the reference point $P(x, y)$ of the moving platform. Accordingly, the constraint equation   of the four-bar linkage is:
\begin{equation}
{\cal C}_2: (x+d\cos(\alpha)-l\cos(\theta_5))^2+
            (y+d\sin(\alpha)-g - l\sin(\theta_5))^2=l^2
\end{equation}
where $\theta_5$  and $\alpha$  are the input and output angles, respectively,  $\left\|{\bf A}_3{\bf B}_3\right\|=\left\|{\bf B}_3{\bf P}\right\|=l$,  $\left\|{\bf C}_3{\bf P}\right\|=d$ and  $\left\|{\bf A}_3{\bf O}\right\|=g$. 
A serial (resp. parallel) singularity is reached whenever $({\bf A}_3{\bf B}_3)$ is aligned with $({\bf B}_3{\bf C}_3)$ (resp. when $({\bf B}_3{\bf C}_3)$ is aligned with $({\bf C}_3{\bf P})$). 
These singularities are defined as follows:
\begin{eqnarray}
\delta_{s3}&:&	(2g \sin(\alpha)-2x \cos(\alpha)-2y\sin(\alpha))d-d^2-x^2-g^2-y^2+4l^2+2yg=0 \nonumber \\
\delta_{p4}&:& 
\begin{array}{l}
g^2+2(l\sin(\alpha)-d\sin(\alpha)-y)g+x^2 \\
(d \cos(\alpha))- 2l \cos(\alpha))x+y^2+(2d\sin(\alpha)-2l\sin(\alpha))y+d^2-2ld=0
\end{array} \nonumber \\
\delta_{p5}&:& 
\begin{array}{l}
g^2-2(l\sin(\alpha)+d\sin(\alpha)+y)g+x^2 + \\
(d \cos(\alpha))+ 2l \cos(\alpha))x+y^2+(2d\sin(\alpha)+2l\sin(\alpha))y+d^2+2ld=0
\end{array} \nonumber
\end{eqnarray}
It is proposed to find those designs for which the platform can be oriented within desired bounds throughout the prescribed workspace. Accordingly, the parameters considered here are the orientation angle $\alpha$ of the moving platform plus only one geometric parameter to handle a two-dimensional parameter space. For the purpose of this study, we choose the distance between the fixed base point $C_3$ and the geometric center of the prescribed workspace: $h=g-f$ and parameter $d$ is set to 1 to have a two dimensional representation of the solution regions. 

\underline{Lam\'e curve prescribed workspace:} From Fig.~\ref{Figure_02}, the smallest value of parameter $l$  is equal to 3. This value is chosen for the four-bar linkage design. The following problem has then to be solved:
\begin{equation}
{\cal P}_{L'}: [h~\alpha]/{\delta}_{p4} \cap {\delta}_{p5} \cap {\delta}_{s3} \cap
\overline{\cal W}_L=\emptyset, h>0
\end{equation}
 
There exist two solution regions, ${\cal R}_{L'1}$   and ${\cal R}_{L'2}$ (Fig.~\ref{Figure_05}), each one being associated with a single working mode and a single assembly mode. These regions describe the orientation ranges as function of parameter $h$, for which the robot can reach the full prescribed workspace without crossing singularities. It is then possible to choose $h$  such that the range of the angular displacement $\alpha$  is greater than a prescribed value.

\begin{figure}[hbtp]
     \begin{center}
 				\includegraphics[height=0.15\textheight]{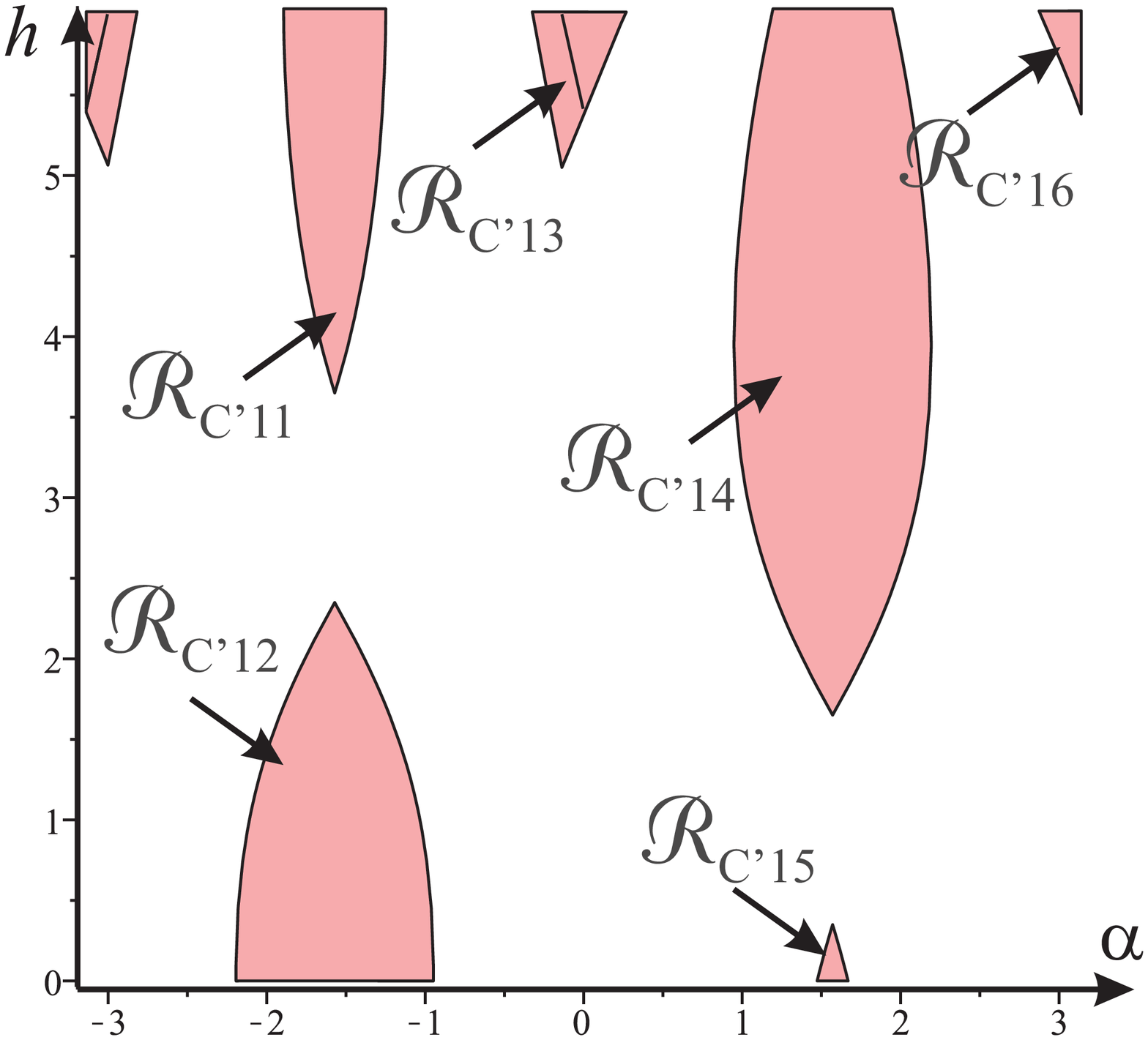}{\small (a)}
 				\includegraphics[height=0.15\textheight]{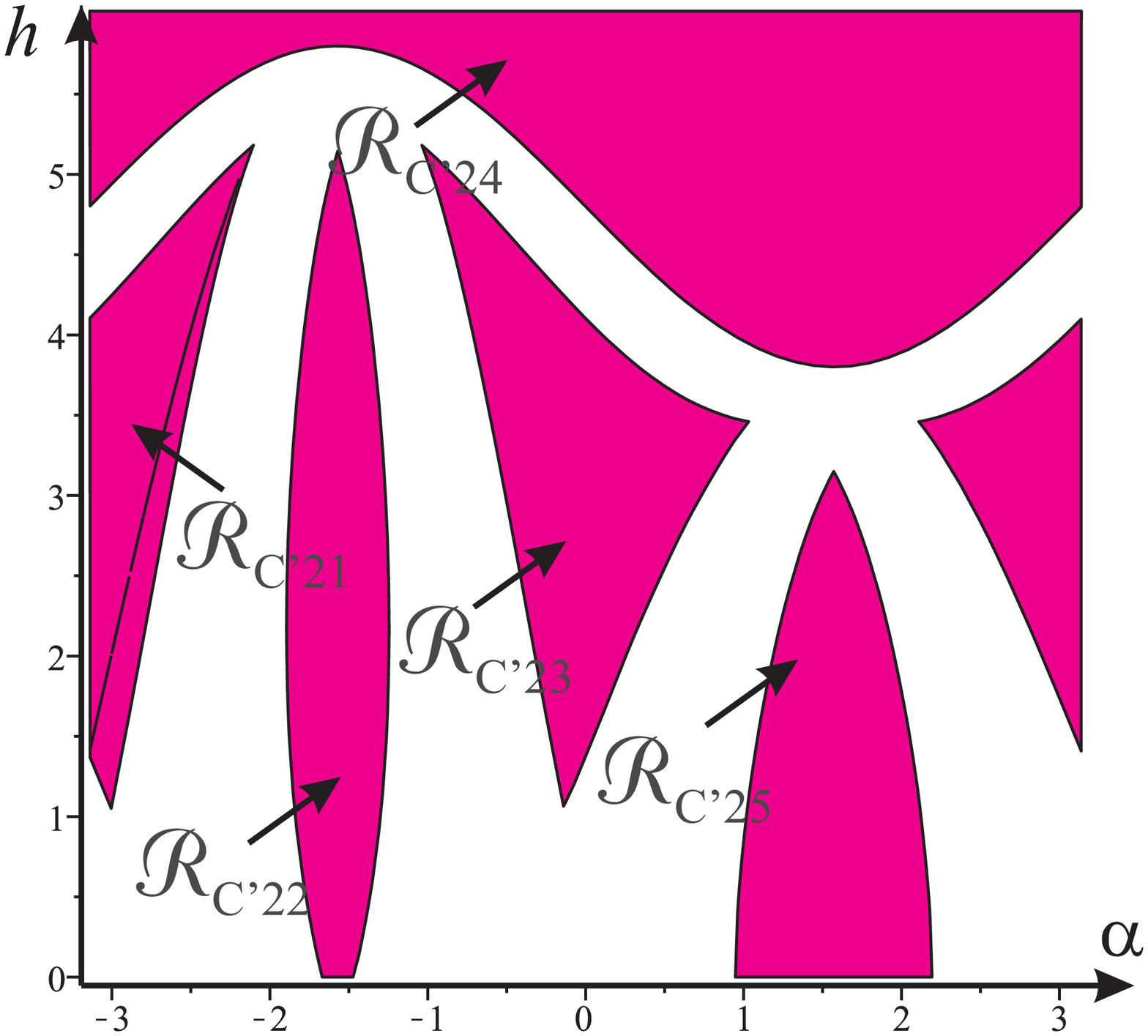}{\small (b)} 
 				\includegraphics[height=0.15\textheight]{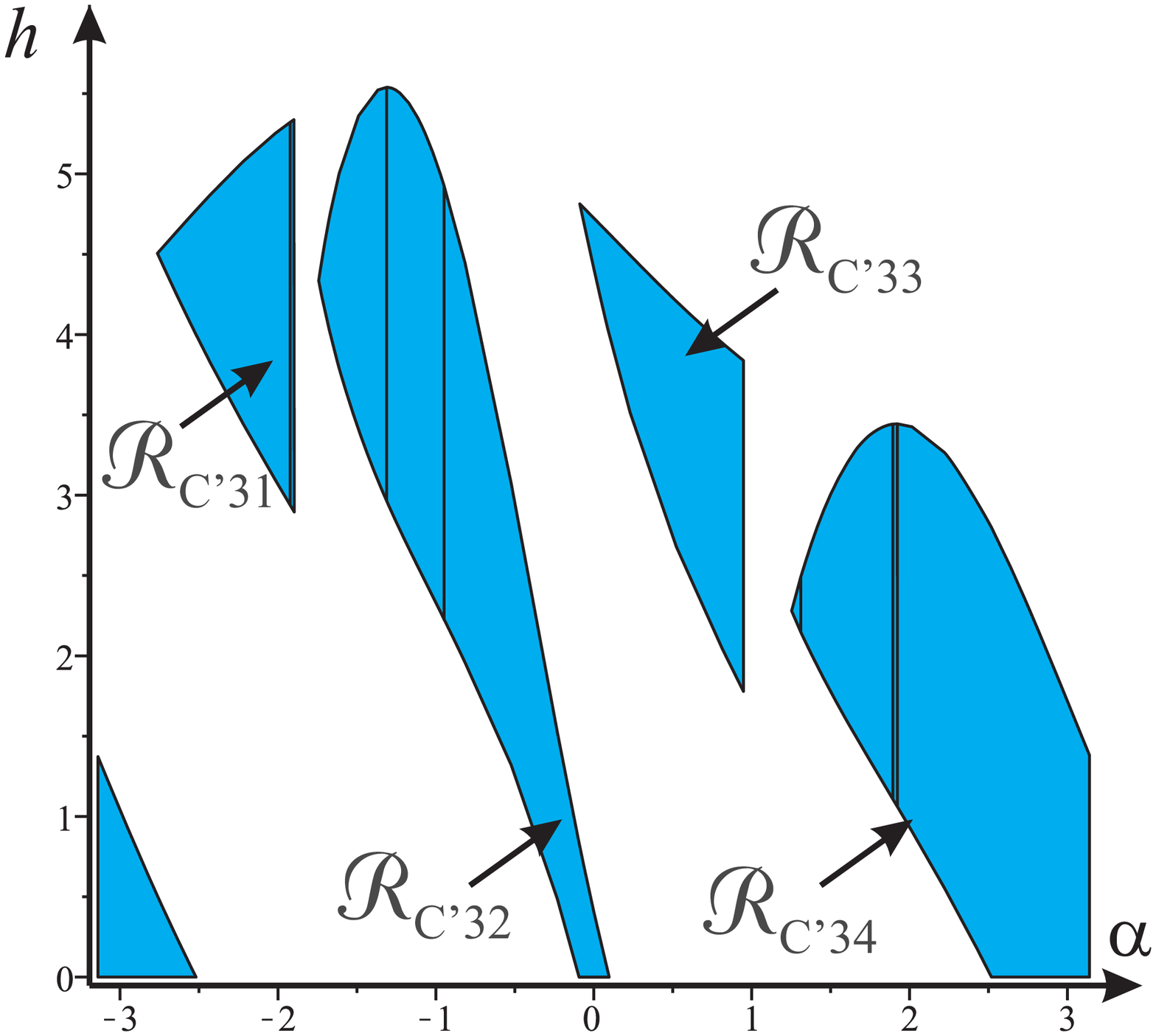}{\small (c)}
        \caption{Solution regions for problems (a) ${\cal P}_{C'1}$, (b)  ${\cal P}_{C'2}$, (c) ${\cal P}_{C'3}$}
        \protect\label{Figure_06}
    \begin{tabular}{cc}
       \begin{minipage}[t]{35 mm}
           \includegraphics[height=0.15\textheight]{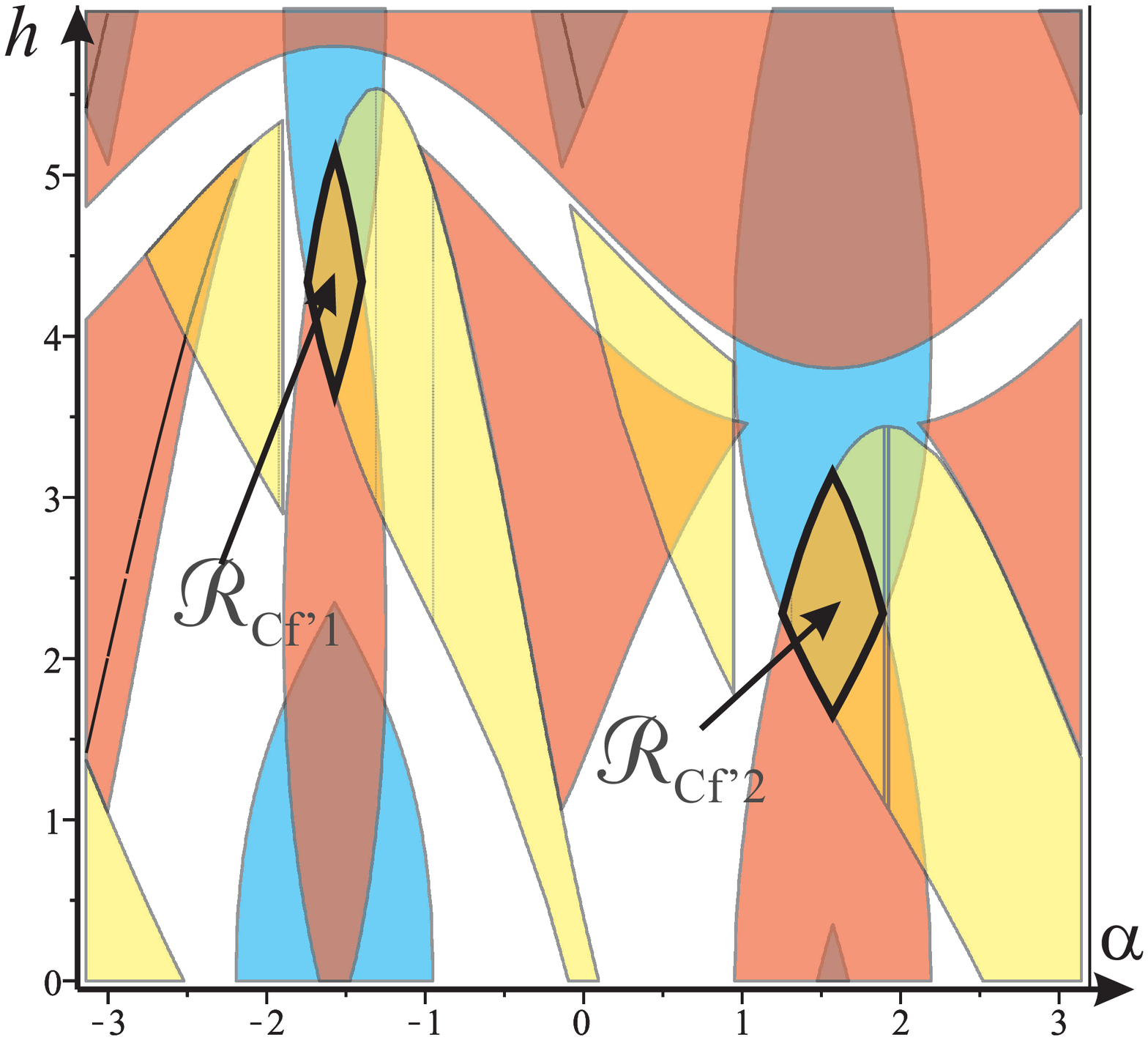}
           \caption{Intersection regions when $l=3.3$}
           \protect\label{Figure_06d}
       \end{minipage} &
       \begin{minipage}[t]{125 mm}
  				\includegraphics[height=0.15\textheight]{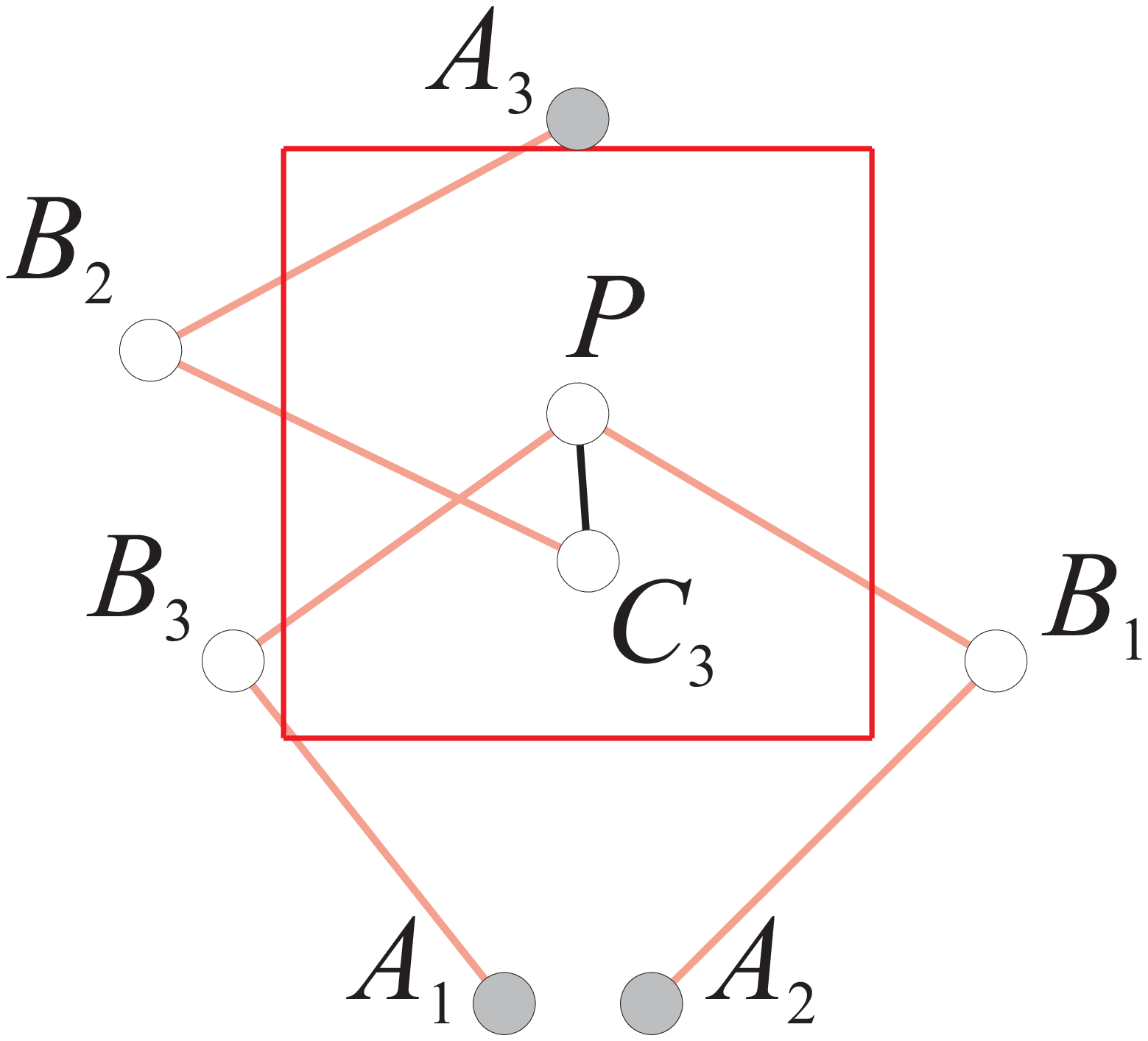}
  				\includegraphics[height=0.15\textheight]{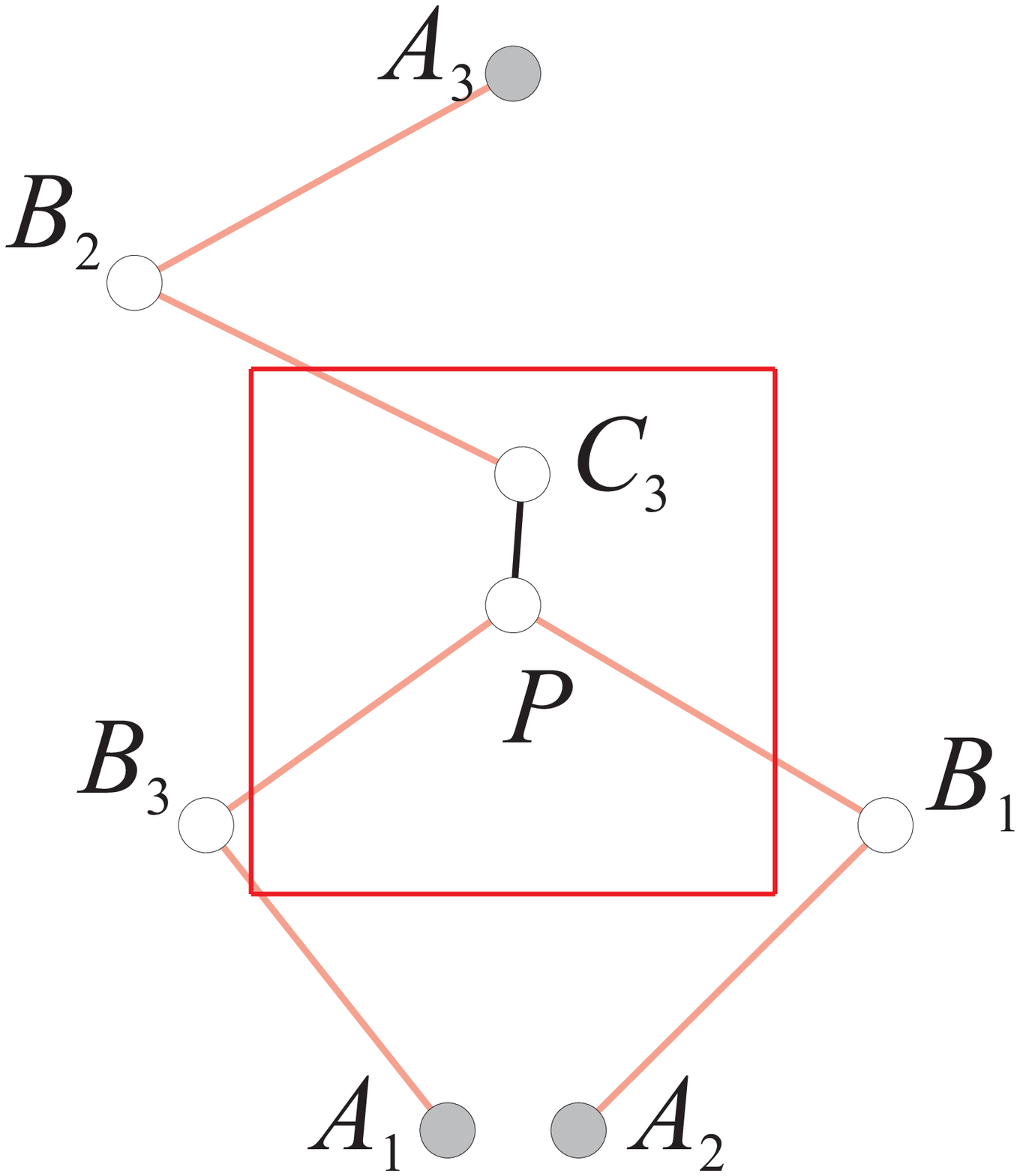}
        \caption{Two examples from the regions ${\cal P}_{c5}$ and ${\cal P}_{c6}$  for (a) \goodbreak $h=4.25$, $\alpha=[-1.717~-1.424]$ and (b) $h=2.2$, $\alpha= [1.306~1.835]$.}
        \protect\label{Figure_07}
      \end{minipage}
    \end{tabular}
    \end{center}
\end{figure}

\underline{Square prescribed workspace:} From Fig.~\ref{Figure_03}, the smallest value of parameter $l$  is equal to $3.3$. This value is chosen for the 4-bar linkage design. The following problems have to be solved:
\begin{equation}
{\cal P}_{C'i}:[h~\alpha]/{\delta}_{p4} \cap {\delta}_{p5} \cap {\delta}_{s3} \cap
\overline{\cal W}_i=\emptyset, h>0, t\in[0~1], i=1, \ldots, 4
\end{equation}
\begin{floatingfigure}[r]{50mm}
 				\includegraphics[height=0.15\textheight]{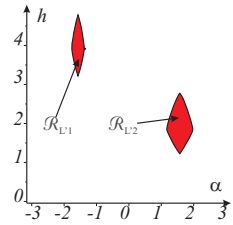}
        \caption{Solution regions of problem ${\cal P}_{L'}$  for a four-bar linkage when  $l=3$}
        \protect\label{Figure_05}
\end{floatingfigure}
The solutions regions of these problems and the intersection regions are shown in Fig.~\ref{Figure_06} and Fig.~\ref{Figure_06d}, respectively. Figure~\ref{Figure_07} depicts two 3-RRR parallel robots obtained for a square regular workspace. The solution obtained in Fig.~\ref{Figure_07}b  is more compact than in Fig.~\ref{Figure_07}a and its angular range interval is greater but the design should take into account the self collisions.
\section{Conclusions}
This paper presented a new design method to determine the feasible set of parameters of parallel manipulators for a prescribed singularity-free regular workspace. Rather than giving a single feasible or optimal solution, this method provides the solution regions in the parameter space. Groebner bases, discriminant varieties and cylindrical algebraic decomposition were used to generate the solution regions. As a result, their boundaries have an exact formulation. Solutions close to the boundaries of these regions correspond to robots for which the prescribed workspace is close to a singularity curve. The prescribed workspace can be defined in a more restrictive way to ensure that the robot will remain far enough from singularities. A solution would be to add a condition relying on some kinetostatic index \cite{Chablat:2004}.
The method was applied to a 3-RRR parallel planar robot with position/orientation decoupled architecture. It can handle other types of translational or decoupled robots but there are some limits that are due to the algebraic tools used. In particular, the number of parameters involved in the elimination process should not be too high. 	 
 
\bibliographystyle{spmpsci}

\end{document}